\documentclass[11pt]{article}
\usepackage{coling2018}
\usepackage{times}
\usepackage{url}
\usepackage{latexsym}
\usepackage{graphicx}
\usepackage{subcaption}
\usepackage{url}
\usepackage{array}

\title{LSTMs with Attention for Aggression Detection}

\author{Nishant Nikhil \\
  IIT Kharagpur \\
  Kharagpur India \\
  {\tt nishantnikhil@iitkgp.ac.in} \\\And
  Ramit Pahwa \\
  IIT Kharagpur  \\
  Kharagpur India \\
  {\tt ramitpahwa123@iitkgp.ac.in} \\ \AND
  Mehul Kumar Nirala \\
  IIT Kharagpur \\
  Kharagpur India \\
  {\tt mehulkumarnirala@iitkgp.ac.in} \\ \And
  Rohan Khilnani \\
  IIT Kharagpur \\
  Kharagpur India \\
  {\tt rkhilnani9@iitkgp.ac.in}}

\date{}

\begin{document}

\maketitle
\begin{abstract}
In this paper, we describe the system submitted for the shared task on Aggression Identification in Facebook posts and comments by the team Nishnik. Previous works demonstrate that LSTMs have achieved remarkable performance in natural language processing tasks. We deploy an LSTM model with an attention unit over it. Our system ranks 6th and 4th in the Hindi subtask for Facebook comments and subtask for generalized social media data respectively. And it ranks 17th and 10th in the corresponding English subtasks.
\end{abstract}

\section{Introduction}
\label{intro}

\blfootnote{
    %
    %
    %
    %
    %
    %
    \hspace{-0.65cm}  
    This work is licensed under a Creative Commons 
    Attribution 4.0 International License.
    License details:
    \url{http://creativecommons.org/licenses/by/4.0/}
}

In recent years, there has been a rapid growth in social media usage. Interactions over the web and social media have seen an exponential increase. While usage of social media helps users stay connected; incidents of aggression, trolling, cyberbullying, flaming, and hate speech are more prevalent now than ever.

Recent works on aggression classification include the use of logistic regression classifier \cite{davidson2017automated}. They create a bunch of hand-crafted features like binary and count indicators for hashtags, lexicon based sentiment scores for each tweet, unigram, bigram, and trigram features. They use two logistic regression models, the first one to reduce dimensionality of the features and the second one to make classification. \newcite{kwok2013locate} train a binary classifier to label tweets into `racist' and  `non-racist'. They deploy Naive Bayes classifier on unigram features.
Neural language model was used in \newcite{djuric2015hate}. First they learn embedding of the text passages using paragraph2vec \cite{Le:2014:DRS:3044805.3045025}. Then, they train a logistic regression classifier over those embeddings to classify into hateful and clean comments. \newcite{schmidt2017survey} surveys the recent development in this field.

The first shared task on aggression identification \cite{trac2018report} was held at the first workshop on Trolling, Aggression and Cyberbullying (TRAC). The goal was to classify social media posts into one of three labels (Overtly aggressive, Covertly aggressive, Non-aggressive).

The major contribution of the work can be summarized as a neural network based model which has LSTM units followed by an attention unit to embed the given social media post and  training a classifier to detect aggression. We discuss our methods in section 2. Section 3 contains the details about the experiments and training data. In Section 4, we discuss the results and Section 5 concludes the paper with closing remarks.

\section{Methodology}

We hypothesize that aggression identification requires processing of the words of a sentence in a sequential manner. The positioning of a particular word at different places can alter the aggressiveness of the sentence. Example: \\

\noindent These aliens are filthy, but they live in a good neighbourhood. (Aggressive) \\ 
These aliens are good, but they live in a filthy neighbourhood. (Less aggressive) \\

Recurrent Neural Networks \cite{mikolov2010recurrent} are good at handling sequential data and have achieved good results in natural language processing tasks. As RNNs share parameters across time, they are capable of conditioning the model on all previous words of a sentence. Although theoretically it is correct that RNNs can retain information from all previous words of a sentence, but practically they fail at handling long-term dependencies. Also, RNNs are prone to the vanishing and exploding gradient problems when dealing with long sequences. Long Short-Term Memory networks \cite{Hochreiter:1997:LSM:1246443.1246450}, a special kind of RNN architecture, were designed to address these problems. 

\subsection{Long Short-Term Memory networks}

LSTMs use special units in addition to standard RNN units. These units include a `memory cell' which can maintain its state for long periods of time. A set of non-linear gates is used to control when information enters the memory(Input gate), when it's outputted (Output gate), and when it's forgotten (Forget gate). The equations for the LSTM memory blocks are given as follows:
\begin{equation}
f_t = \sigma_g(W_{f} x_t + U_{f} h_{t-1} + b_f) \\
\end{equation}
\begin{equation}
i_t = \sigma_g(W_{i} x_t + U_{i} h_{t-1} + b_i) \\
\end{equation}
\begin{equation}
o_t = \sigma_g(W_{o} x_t + U_{o} h_{t-1} + b_o) \\
\end{equation}
\begin{equation}
c_t = f_t \circ c_{t-1} + i_t \circ \sigma_c(W_{c} x_t + U_{c} h_{t-1} + b_c) \\
\end{equation}
\begin{equation}
h_t = o_t \circ \sigma_h(c_t)
\end{equation}

In these equations, $x_t$ is the input vector to the LSTM unit, $f_t$ is the forget gate's activation vector, $i_t$ is the input gate's activation vector, $o_t$ is the output gate's activation vector, $h_t$ is the output vector of the LSTM unit and $c_t$ is the cell state vector. $w, u, B $ are the parameters of weight matrices and bias vectors which are learned during the training.

\subsection{Attention}

Here, the attention module is inspired by \cite{DBLP:journals/corr/BahdanauCB14}. We deploy it after the LSTM unit. It helps the model decide the importance of each word for the classification task. It scales the representation of the words by a learned weighing factor, as determined by these equations:
\begin{equation}
e_t = h_t w_a \\
\end{equation}
\begin{equation}
a_t = \frac{exp(e_t)}{\sum_{i=1}^{T} exp(e_i)} \\
\end{equation}
\begin{equation}
v = \sum_{i=1}^{T} a_i h_i
\end{equation}

In these equations, $h_t$ is the hidden representation of a word at a time step $t$, $w_a$ is the weight matrix for the attention layer, $a_t$ is the attention score for the word at time $t$, and $v$ is the final representation of the sentence obtained by taking a weighted summation over all time steps.

\section{Experiments}
\subsection{Datasets}
The training datasets for the English and Hindi sub-tasks are constructed by 11,999 and 12,000 Facebook posts and comments, respectively. The testing set includes 3,001 and 3,000 respectively. These are collected and manually annotated by the organizers. Most of the datum has an id and is classified into one of the three classes: OAG (Overtly aggressive), CAG (Covertly aggressive), NAG (Non-aggressive). The distribution of the classes in the training dataset for English and Hindi are shown in Table 1. The organizers released a modified version of the data where they remove the rows without specific id. As the id is not important for prediction, we decided to work on the initially released dataset. The data collection methods used to compile the dataset for the shared task are described in \newcite{trac2018dataset}. 

\begin{table*}[h]
\center
\begin{tabular}{|l|llll|}
\hline
Class & Train (English) & Test (English) & Train (Hindi) & Test (Hindi) \\
\hline
Non-aggressive & 5,051 & 1,233 & 2,275 & 538\\
Covertly aggressive & 4,240 & 1,057 & 4,869 & 1,246\\
Overtly aggressive & 2,708 & 711 & 4,856 & 1,217\\
\hline
\end{tabular}
\caption{Class distribution in train and test sets}
\label{tab:class_desc}
\end{table*}

\subsection{Preprocessing}
Before feeding the Facebook comments to the LSTM classifier, we performed the following operations on the text:
\begin{enumerate}
\item We used the ekphrasis toolkit \cite{S17-2126} for normalizing the occurrence of the following in the comments: URL, E-mail, percent, money, phone, user, time, date, and number. For example, URLs are replaced by \textless url\textgreater, and all occurrences of @someone are replaced by \textless user\textgreater.
\item We then passed the normalized text through the Social tokenizer. Unlike normal tokenizers, the Social tokenizer is specifically aimed at the unstructured social media content. It understands and parses complex emoticons, emojis and other unstructured expressions like dates, times, phone numbers etc.
\item Then, we removed the punctuations and used ekphrasis's inbuilt spell corrector on the text.
\item Lastly, we used NLTK's WordNet lemmatizer \cite{Loper:2002:NNL:1118108.1118117} to lemmatize the words to their roots.
\end{enumerate}

\subsection{Parameters}

Our model uses an embedding layer of 100 dimensions to project each word into a vector space. We place a dropout \cite{Srivastava:2014:DSW:2627435.2670313} layer after this. To capture the context of the words passed from the dropout layer we use an LSTM layer having 100 hidden dimensions. As the LSTM cells already have non-linear activation functions, it helps the model capture non-linear semantics from the data. The output from the LSTM is then passed through an attention module. The attention module helps the model determine which word to give more importance to. The weighted output from attention module is passed through a fully-connected layer. To get the probabilities of each class, softmax function is applied to the output. We use the cross-entropy function to calculate the loss between the predicted and the target value. Adam optimizer is used with a learning rate of 0.001 to learn the weights of the model. The dropout rate was either 0.2 or 0.3 and is discussed in the Results section.

Although many machine learning classifiers like Naive Bayes, Decision Tree, Support Vector Machine or Random Forest could be used as a baseline classifier for this task. Due to constraint of time we have only used a Random Forest classifier. We train the classifier on a set of hand-crafted features. The features used are as follows:
\begin{enumerate}
\item Number of words with positive sentiment.
\item Number of words with negative sentiment.
\item Number of punctuations.
\item Total number of words.
\item Inverse of the 2nd feature.
\item Natural logarithm of the 2nd feature.
\end{enumerate}

We use the lists made available by \newcite{Hu:2004:MSC:1014052.1014073} for extracting positive and negative words. 

\section{Results}
\label{sec:results}

Due to a mistake on our side, we first submitted a model which considered only the first 45 words of the post/comment and used a dropout rate of 0.2, we denote this model as Eng-A in the tables. In Eng-B, we use dropout rate of 0.3 and considered all the words. RF baseline is the random forest classifier based model baseline of hand-crafted features.
\begin{table*}[h]
\center
\begin{tabular}{|ll|}
\hline
\bf System & \bf F1 (weighted) \\ 
\hline
Random Baseline & 0.3535 \\
\hline
EF-A & 0.5533 \\
EF-B & \textbf{0.5746} \\
\hline
\end{tabular}
\caption{Results for the English (Facebook) task.}
\label{tab:results-EN-FB-open}
\end{table*}

\begin{table*}[h]
\center
\begin{tabular}{|ll|}
\hline
\bf System & \bf F1 (weighted) \\ 
\hline
Random Baseline & 0.3477 \\
\hline
RF Baseline & 0.3888 \\
Eng-A & 0.5304 \\
Eng-B & \textbf{0.5548} \\
\hline
\end{tabular}
\caption{Results for the English (Social Media) task.}
\label{tab:results-EN-TW-open}
\end{table*}

For both the Hindi sub-tasks, we used the LSTM classifier with dropout probability of 0.3. We denote the model as Hi-A in the tables.

\begin{table*}[!h]
\center
\begin{tabular}{|ll|}
\hline
\bf System & \bf F1 (weighted) \\ 
\hline
Random Baseline & 0.3571 \\
\hline
Hi-A & \textbf{0.6032} \\
\hline
\end{tabular}
\caption{Results for the Hindi (Facebook) task.}
\label{tab:results-HI-FB-open}
\end{table*}

\begin{table*}[!h]
\center
\begin{tabular}{|ll|}
\hline
\bf System & \bf F1 (weighted) \\ 
\hline
Random Baseline & 0.3206 \\
\hline
Hi-A & \textbf{0.4703} \\
\hline
\end{tabular}
\caption{Results for the Hindi (Social Media) task.}
\label{tab:results-HI-TW-open}
\end{table*}

Looking at the confusion matrices of the English subtasks, it is clear that the model is performing well at classifying the non-aggressive comments from the aggressive or covertly aggressive comments. But it performs poorly and classifies a lot of over-aggressive and non-aggressive comments to covertly aggressive. The results in \newcite{malmasi2018challenges} also convey the same message.

\begin{figure*}[!h]
\centering
\includegraphics[width=0.5\textwidth]{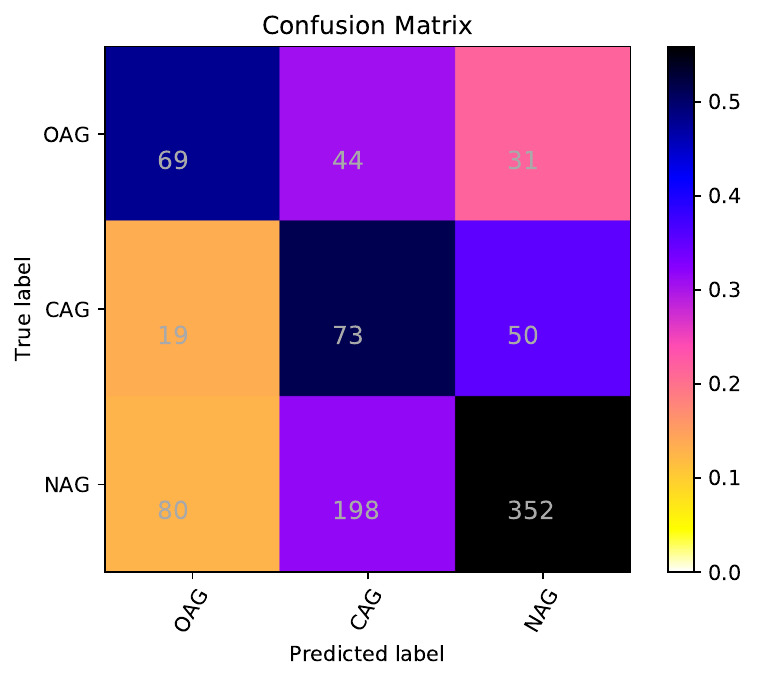}
\caption{Confusion matrix for English (Facebook) task.}
\label{fig:1}
\end{figure*}

\section{Conclusion}

In this paper, we present an LSTM network with an attention based classifier for aggression detection. It gives competitive results while relying only on the dataset provided. The performance reported in this paper could be further boosted by utilizing transfer learning methods from larger datasets, like using pre-trained word embeddings. Furthermore, the model tends to over-fit on the training data. Better generalization techniques, like the use of an increased dropout rate, might help in increasing the performance of the model.


\bibliography{trac}
\bibliographystyle{acl}

\end{document}